%% file: main.tex
\documentclass[letterpaper]{article}

\usepackage{iccc}
\usepackage{graphicx}
\usepackage{subcaption}
\usepackage{natbib}
\usepackage{booktabs}

\newcommand{\point}[1]{\par\noindent \raisebox{.3ex}{$\bullet$}\hskip 0.05in #1}

\title{\textbf{Coherence-Oriented Dream Scene Visualisation} }

\author{Azra Açıl and Simon Colton\\ \\ School of Electronic Engineering and Computer Science, Queen Mary University of London, UK}

\begin{document}

\maketitle

\begin{abstract}
Dreams can be emotionally intense but difficult to communicate. We describe the Dream Scene Visualiser (DSV) system which turns written dream descriptions into a temporal sequence of four panel images visualising the dream. This starts with a large language model prompted to split a dream description into four chronological parts. Then a text-to-image model produces images for each part with visual coherence maintained across the sequence, and DSV regenerates any image not suitably matching the text. We evaluate DSV over 50 visualisations from dream descriptions in DreamBank, and report quality, fidelity and coherence results via objective measures employing the CLIP, DINOv2 and Qwen2-VL vision-language models.
\end{abstract}

\section{Introduction}

Dreams are one of the most universal forms of human experience, yet communicating them to other people can be surprisingly hard. They are often deeply personal, emotionally vivid, and sometimes lack narrative logic, making them difficult to convey through words alone. Visualising dreams can have significant value. Dream journaling has been shown to support self-awareness and memory consolidation~\citep{schredl2002dreamdiary}. Visual dream records could also support clinical practice. For instance, in psychotherapy, patients often struggle to verbalise the atmosphere of a dream, and a shared visual representation could give the therapist and patient a common reference point for mood, imagery, and emotional tone. This is particularly relevant for trauma-related dreams, where Image Rehearsal Therapy uses patient-generated imagery therapeutically~\citep{krakow2001imagery}, and similar to~\citet{cheatley2020songwriting} using co-creative songwriting to support people who lack artistic expertise, during bereavement.

Existing tools have limitations for dream visualisation. In particular, text-to-image models can produce a striking single image but not a sequence with its own mood, pacing, and emotional progression. Story-visualisation models such as those in StoryDALL-E~\citep{maharana2022storydalle} and AR-LDM~\citep{pan2023arldm} handle image sequences. However, they were trained on paired text-image datasets largely consisting of everyday imagery. Hence they tend to smooth out unusual features that make dreams distinctive. As there is no dataset of dream visualisations matched to text, a different approach is needed.

We present the Dream Scene Visualiser (DSV), which turns a written dream description into a visualisation of the dream in four panels, each containing an image and textual information. In addition to image sophistication and fidelity to the dream narrative and mood, a key consideration for the process is that coherence is maintained over the whole visualisation and from panel to panel. In the next section, we describe how narrative coherence is maintained using the Qwen2.5(14B) large language model (LLM)~\citep{yang2024qwen25} to break the dream description into parts. Moreover, visual coherence is maintained by seeding the generation of panel images 2, 3 and 4 with the previous image, using the SDXL text-2-image generator~\citep{podell2023sdxl}. Finally, for semantic coherence, a feedback loop using the vision-language model CLIP(ViT-B/32)~\citep{radford2021clip}, measures if a panel image suitably matches the description, regenerating any image which does not.

We evaluate the system on 50 dream descriptions from the DreamBank repository~\citep{domhoff2008dreambank}. The evaluation protocol involves assessing text/image alignment using CLIP and panel-to-panel visual similarity, using DINOv2~\citep{oquab2024dinov2}, a self-supervised vision model. We also use the Qwen2-VL(7B)~\citep{wang2024qwen2vl} vision-language model (VLM) as an independent judge evaluating seven dimensions of quality in the dream visualisations.

\section{The DSV Pipeline}

\begin{figure*}[t]
\centering
\includegraphics[width=0.95\textwidth]{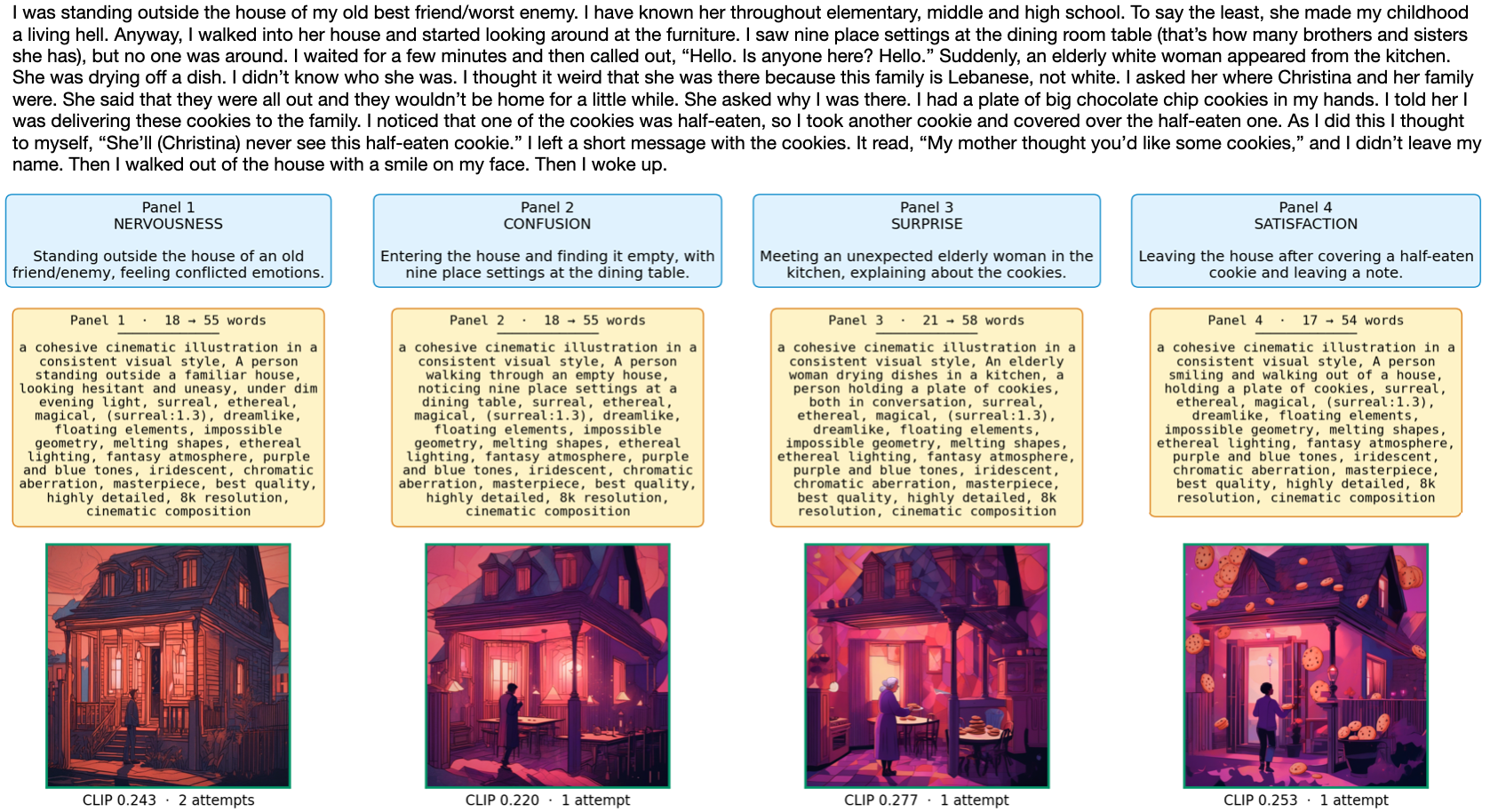}
\vskip -0.05in
\caption{Pipeline trace for DreamBank example \texttt{DB-b533e9bb}: (a) original dream text (b) decomposition into four short panel texts and dominant emotion label (c) enriched panel texts for use as prompts in SDXL (d) generated panel images after post-processing, given with feedback loop repetitions and CLIP cosine similarities to the texts.}
\label{fig:pipeline-trace}
\vskip -0.1in
\end{figure*}

A dream description is first annotated with two attributes before entering the pipeline: a complexity tier between 1 and 5, and a dominant style label of either \textit{surreal}, \textit{nightmare}, \textit{peaceful} or \textit{fantastical}. Annotation is performed by Qwen2.5(14B Instruct) using a prompt based on the dream coding system of ~\citet{halldream1966}. The full prompt is given in appendix A. Figure~\ref{fig:pipeline-trace} overleaf shows a trace of the DSV pipeline applied to a representative DreamBank report. The pipeline processes are:

\vskip 0.05in
\point{\textbf{Panel decomposition.} For narrative coherence, a dream is first split into four chronological panel texts. This is achieved by prompting the Qwen2.5(14B) LLM with temperature 0, using the prompt given in appendix B. This produces (a) a short description for each panel (b) a dominant emotion label (c) an SDXL-ready visual prompt and (d) a negative SDXL prompt.}

\vskip 0.05in
\point{\textbf{Prompt enrichment.} The LLM-generated prompts are short  and do not carry enough stylistic signal for SDXL to produce a coherent visual style across the four panels. We therefore enrich each prompt by prepending the style label annotation, then appending style-specific positive tokens, emotion-specific tokens and a generic quality booster. This stage brings the LLM prompt from around 18 words to around 55 words per image, on average in the example of figure \ref{fig:pipeline-trace}. Details of how token dictionaries are used for enriching positive and negative SDXL prompts are given in appendix C.}

\vskip 0.05in
\point{\textbf{Image generation.}
For visual coherence, all images are generated with SDXL 1.0 at $768 \times 768$ pixel resolution. SDXL generates images to match a positive prompt and avoid matching a negative prompt. Panel 1 is generated from the first panel positive and negative prompts, but panels 2 to 4 use \texttt{img2img} chaining, conditioned on the previous panel image. This means that, along with the positive and negative prompts, SDXL is forced to use the previous image, with a weighting of 0.3/0.7 respectively. These weights were determined empirically: \texttt{img2img} strengths below 0.5 collapse panels onto near-duplicates of panel 1, while strengths above 0.8 lose character and setting continuity. We found that strength 0.7 preserves recognisable continuity, while allowing each panel to introduce its own content. This mechanism works best when the dream stays within a single setting. As an example, in figure \ref{fig:pipeline-trace}, the dark house, the dreamer in silhouette, and the red-purple palette all persist across all four panels. }

\vskip 0.05in
\point{\textbf{CLIP feedback loop.}
For semantic coherence, after each image is generated, we use CLIP to measure the cosine similarity between the image and the text description of each panel, with higher values indicating closer agreement between the two. If the alignment score falls below a calibrated threshold of 0.228, the panel is regenerated with a prompt revised via emphasis changes and token additions such as ``sharp focus'', with up to three attempts allowed in total. In the example of figure \ref{fig:pipeline-trace}, the loop fired once on panel 1, as the first attempt fell below the threshold, and the revised prompt on attempt 2 was accepted with CLIP cosine similarity of 0.243.}

\vskip 0.05in
\point{\textbf{Post-processing.}
Each of the four images receives a lightweight cosmetic pass, bringing contrast to 1.15 and brightness to 1.05. Following this, a Gaussian blur with $\sigma = 0.4$, and a radial vignette at intensity 0.25 are applied. We have found such post-processing helps unify the four panels visually for dream-journal presentation.}

\vskip 0.05in
\par\noindent
Figure 1 shows a \emph{surreal} example, and appendix F contains additional example visualisations in the \emph{nightmare}, \emph{fantastical}, and \emph{peaceful} categories.

\section{Evaluation}

Fixing the panel count for visualisations at four keeps the output format consistent across dreams of different length. This has enabled an evaluation of DSV on DreamBank reports with three complementary signals: text-image alignment, panel-to-panel visual coherence, and a seven-dimension vision-language judge.

We sampled 50 dream reports from DreamBank, drawn from eight dreamer series spanning everyday, lucid, anxiety-laden, and long-narrative dream types. The dream reports were annotated using Qwen with a dream style as follows: \emph{nightmare}: 18 dreams, \emph{fantastical}: 6, \emph{surreal}: 14 and \emph{peaceful}: 12. The word counts for the enriched prompts range from 52 to 412, with a median of 194. CLIP ViT-B/32 was used to measure image-to-text alignment between each generated panel and its enriched description; we report per-dream averages. We further used the DINOv2 model to measure visual similarity between adjacent panels, with a high score indicating that consecutive panels share a character, palette, and setting, which is the form of coherence \texttt{img2img} chaining is designed to produce.

To further evaluate the dream visualisations, we implemented a \emph{judge} using the Qwen2-VL(7B) vision-language model. Full details of how the model is prompted in order to pass judgement on the dream visualisations are given in appendix D. In overview, the judge is called twice per dream. Firstly, it is employed once per panel to judge four aesthetic qualities of the individual images, namely: \textit{aesthetic quality}, \textit{character presence}, \textit{dreamlike quality} and \textit{text faithfulness}. Secondly, it is given all four panel images in one image, and asked to judge three per-sequence dimensions, namely: \textit{style consistency}, \textit{narrative coherence} and \textit{emotional progression}. The judge is prompted to score each image/sequence quality on a 1 to 5 Likert scale, with 1 indicating low quality and 5 indicating high quality.

Across 50 dreams, using the full pipeline produces mean CLIP cosine similarity for image/text pairs of $0.250 \pm 0.020$ and mean adjacent-panel DINOv2 coherence of $0.623 \pm 0.143$. Shuffled-baseline controls confirm that both signals are above chance: matched CLIP=$0.252$ vs shuffled=$0.201$ ($p=6 \times 10^{-25}$, paired Wilcoxon), chained DINOv2=$0.623$ vs random 4-tuples=$0.182$ ($p = 1.4 \times 10^{-27}$, Mann--Whitney U). The two metrics are weakly correlated ($r=+0.24$), confirming they measure different aspects of coherence. 

The scores assigned by the judge are given in table~\ref{tab:judge}. These indicate that high style consistency and aesthetic quality (both above~4/5) are achieved, with moderate text faithfulness (3.2), but lower scores for sequence-level narrative and emotional coherence. This is consistent with the production of stylistically strong, visually continuous imagery, but only partially captures narrative detail. CLIP correlates moderately with the judge's text-faithfulness score ($r=+0.38, p=0.007$), supporting CLIP usage as a coarse alignment signal rather than a stand-in for perceptual quality. We found that adjacent DINOv2 coherence decreases along the chain (from 0.66 at P1$\to$P2 to 0.58 at P3$\to$P4), reflecting accumulated visual drift from repeated \texttt{img2img} conditioning.

\begin{table}[t]
\centering
\small
\begin{tabular}{|l||c|c|}
\hline
\textbf{Judge dimension} & \textbf{Mean} & \textbf{Std} \\
\hline
\emph{Per-panel (N=200)} & & \\
Aesthetic quality         & 4.41 & 0.62 \\
Character presence        & 3.81 & 1.33 \\
Dreamlike quality         & 3.67 & 0.72 \\
Text faithfulness         & 3.21 & 1.11 \\
\hline
\emph{Per-sequence (N=50)} & & \\
Style consistency         & 4.92 & 0.57 \\
Narrative coherence       & 2.68 & 1.17 \\
Emotional progression     & 2.30 & 1.33 \\
\hline
\end{tabular}
\caption{Judge ratings on a 1 to 5 Likert scale. Per-panel dimensions averaged over 200 generated panels; per-sequence dimensions are single ratings per dream.}
\vskip -0.1in
\label{tab:judge}
\end{table}

\vskip 0.1in
\point{\textbf{A subjective evaluation.} 
In a small pilot study (N=10), subjects rated visualisations for their own dream descriptions, using 1 to 5 Likert evaluations covering style, emotional palette, creative detail, narrative accuracy, object fidelity, and transitions. We found that the ratings align with the automated judge: style application scored highest (av. 4.40), followed by emotional palette (4.20) and creative detail (4.00), while object fidelity (3.60) and transitions (3.50) scored lowest.
}

\begin{table}[b]
\centering
\small
\begin{tabular}{|l||c|c|c|c|}
\hline
\textbf{Condition} & \textbf{CLIP} & \textbf{$\Delta$} & \textbf{$p$} & \textbf{Sig.} \\
\hline
A. Full pipeline     & 0.250 & ---      & ---   & ---  \\
B. No prompt eng.    & 0.260 & +0.010   & 0.002 & s.   \\
C. No CLIP loop      & 0.249 & $-$0.001 & 0.625 & n.s. \\
D. No img2img        & 0.265 & +0.016   & 0.014 & n.s. \\
E. Baseline          & 0.265 & +0.015   & 0.004 & s.   \\
\hline
\end{tabular}
\caption{Ablation over 10 dreams: CLIP: per-dream mean; $\Delta$: difference from condition~A; $p$-values and significance with s.\ = significant and n.s.\ = not significant.}
\label{tab:ablation}
\end{table}

\vskip 0.1in
\point{\textbf{An ablation study.}
We ablate each pipeline stage on a held-out subset of 10 dream descriptions. Starting from the full pipeline (A), we remove prompt engineering (B), the CLIP feedback loop (C), \texttt{img2img} chaining (D), and finally reduce the system to SDXL on unprocessed LLM prompts (E, baseline). Each ablation is tested against A with a two-sided Wilcoxon signed-rank test with Bonferroni correction ($\alpha = 0.0125$). The results are given in table~\ref{tab:ablation}. We see that removing prompt engineering (B) or returning to baseline (E) both significantly increase CLIP alignment relative to the full pipeline. Removing \texttt{img2img} (D) shows a marginal effect that does not reach significance after correction. We also see that removing the CLIP loop (C) shows no significant change, which is because the feedback loop rarely caused re-generations, in fact it happened for only 3\% of panels.}

\vskip 0.1in
\point{\textbf{LoRA fine-tuning.}
To investigate whether the process could be improved with appeal to more dream-like imagery, we fine-tuned a LoRA of rank 16 on SDXL's UNet using 50 surreal artworks found on Pinterest. LoRA~\citep{hu2021lora} is a parameter-efficient fine-tuning method that adapts pre-trained models without modifying their weights. On 5 surreal dreams sampled from the evaluation set, there was a mean CLIP cosine similarity improvement from 0.280 to 0.290, with 4 out of 5 dreams being improved. On generic prompts, however, the CLIP score dropped by 0.009. We found that the effect is content-conditional, i.e., the fine-tuning helps where the base model's priors are weakest and hinders where they are already strong.}

\section{Discussion}

DSV produces stylistically strong and visually coherent sequences, with judge scores for style consistency of 4.92/5 and aesthetic value of 4.41/5. However, the full pipeline approach scores significantly lower for CLIP cosine similarity than a minimal baseline ($p < 0.005$). Moreover, CLIP scores drop when we add our coherence-oriented components, as shown in table~\ref{tab:ablation}. We also see that DINOv2 coherence decays along the \texttt{img2img} chain, and the judge rates the sequences highly on perceptual quality. So it seems that our design choices for DSV improve the output. However, CLIP indicates the opposite. This pattern is partly explained by SDXL's 77-token text encoder limit: enriched prompts exceed this limit, and SDXL silently drops the overflow, which contain the emotion and narrative tokens at the end of our template. This helps to explain the judge's weak per-sequence scores (narrative 2.68, emotional 2.30) alongside strong per-panel ones (all above 3.2). CLIP cannot detect this gap because it scores against the short panel description, not the truncated enriched prompt.

A natural concern is whether the judge simply rewards high quality images. Three observations argue against such bias: it discriminates between dimensions (style 4.92 vs emotional progression 2.30), its text-faithfulness correlates with CLIP ($r = +0.38, p = 0.007$), and the low narrative scores have a mechanical explanation in the silent truncation issue mentioned above. Nevertheless, a VLM-based judge is a proxy for human judgement, not a replacement.

Two dreams are provided in appendix E which illustrate the failure modes of single-metric evaluation. DreamBank dream \texttt{DB-3c0442ac} receives above-mean CLIP (0.263) but the judge scores it as 2.5/5 on text faithfulness and 1/5 on emotional progression: CLIP rewards a coherent palette while missing character drift, a false positive. Dream \texttt{DB-dad219b7} receives a below-mean CLIP score of 0.231, but 5/5 on aesthetic quality and style consistency. Here, stylisation is penalised by CLIP even though the content is clearly there, which is a false negative. No single metric separates these two cases, but the multi-signal approach does. 

Our approach has several limitations. While a pilot human study ($N = 10$) supports the judge's main trends, its sample is too small to stand alone, and VLM judges more broadly are known to exhibit verbosity and self-enhancement biases~\citep{zheng2024mtbench}. A larger human study therefore remains the strongest test of the trade-off reported here. The complexity annotation was not used in the pipeline, and we found that tier-1 and tier-5 complexity categories are under-sampled ($n = 1$ out of $50$ for each). Moreover, the LoRA ablation covers only five surreal dreams, and the CLIP feedback loop fires on just 3\% of panels, functioning as a safety net rather than an active quality gate.

\section{Related Work}


Two recent systems address dream content directly. DreamStory~\citep{he2024dreamstory} uses an LLM as a ``story director'' paired with a specialised diffusion model for character consistency, but targets open-domain fiction such as fairytales. DSV differs by targeting dream reports specifically, adding a CLIP-based self-correction loop, and using a multi-dimensional evaluation protocol. DreamLLM-3D~\citep{liu2024dreamllm3d} takes a different path, displaying dream entities as animated 3D point clouds inside an art installation. In other directions, video dream visualisations from fMRI scans are beginning to be explored \citep{fMRI}. Recent work on using large language models as evaluative judges~\citep{franceschelli2024creative} offers a flexible alternative to traditional metrics, reasoning about creative quality beyond surface-level matching. Our evaluation builds on this direction by combining a vision-language judge with CLIP-based and DINOv2-based signals, motivating the multi-dimensional protocol we use in place of any single metric.

\section{Conclusion and Future Work}

We presented the Dream Scene Visualiser, a coherence-oriented system that turns written dream descriptions into four-image visualisations. On 50 DreamBank dreams, DSV produces stylistically strong and visually continuous sequences (judge style 4.92/5, aesthetic 4.41/5) while scoring lower on CLIP than a minimal baseline ($p<0.005$). We traced this style-fidelity trade-off to silent prompt truncation in SDXL's text encoder, and argued that CLIP alone is insufficient to evaluate dream visualisations. Directions for future work include broader LoRA fine-tuning across all four dream styles and addressing SDXL's 77-token limit, by exploring prompt-engineering strategies that keep prompts within the budget while still including emotion and narrative anchors. We also plan to use the complexity tier annotation to support dynamic panel counts, replacing the current fixed four-panel structure. We hope to eventually trial dream visualisation tools in psychotherapy and other counselling contexts.

\section*{Acknowledgments}
We thank the participants of our pilot study for their time and the maintainers of DreamBank for making dream descriptions 
publicly available. We also thank the anonymous reviewers for their helpful feedback.

\input{bibliography}
\newpage
\input{appendixA}
\input{appendixB}
\input{appendixC}
\input{appendixD}
\input{appendixE}
\input{appendixF}

\end{document}

%% file: appendixA.tex
\section*{Appendix A:\\Annotation LLM Prompt}

Prior to annotation, DreamBank header markers (e.g.\ ``\texttt{\#028 (05/13/2000)}'') are stripped from each dream text. The dream text is then appended to the following prompt for the Qwen2.5(14B Instruct) LLM  (with temperature set to 0), and the JSON output is processed to annotate the dream text.

\vskip 0.1in
\hrule
\vskip 0.05in
\begin{quote}\small
You are a dream researcher trained in the Hall \& Van 
de Castle coding system, annotating dreams for visual 
art generation.

For the following dream, assess two dimensions.

\textbf{1. Complexity (integer, 1--5):}
\begin{itemize}\itemsep0pt
  \item \textbf{1} -- short report; single setting; no 
  bizarre elements
  \item \textbf{2} -- single setting; one action thread; 
  minor anomalies
  \item \textbf{3} -- multiple settings OR clear 
  anomalies (time distortion, impossible objects)
  \item \textbf{4} -- narrative arc across 3+ settings; 
  bizarre transformations; dream-logic
  \item \textbf{5} -- highly complex; multiple 
  transformations; composite characters/settings; 
  impossible physics
\end{itemize}

\textbf{2. Dominant style} --- judge by emotional tone 
and visual atmosphere, not by whether the setting is 
everyday. Apply the first criterion that fits:
\begin{itemize}\itemsep0pt
  \item \textbf{nightmare} -- fear, threat, being 
  chased, trapped, unsettling figures, violence, 
  property damage, police pursuit, anxiety, distress, 
  or dark atmosphere.
  \item \textbf{fantastical} -- magical or 
  impossible-but-positive elements: flying 
  (leisurely, not falling), magic, mythical 
  creatures, cosmic scale, enchanted objects, walking 
  through portals with wonder.
  \item \textbf{surreal} -- reality bending in ways 
  that are strange but not clearly positive or 
  negative: melting shapes, impossible geometry, 
  objects transforming, composite characters, 
  dream-logic without clear emotional valence.
  \item \textbf{peaceful} -- default for everything 
  else: calm, warm, nostalgic, tender, joyful, proud, 
  loving, or restful, or emotionally flat everyday 
  content (having coffee, walking home, ordinary 
  conversations).
\end{itemize}

Respond only with valid JSON: 
\texttt{\{"complexity": <1-5>, "dominant\_style": 
"<label>", "reasoning": "<one-sentence rationale>"\}}
\end{quote}
\vskip 0.05in
\hrule

%% file: appendixB.tex
\section*{Appendix B:\\Panel Decomposition LLM Prompt}
\label{app:llm-prompt}

Qwen2.5(14B Instruct) with temperature 0 is used to generate four panel description texts from the overall dream text. The dream text is appended to the prompt given below. In the prompt, the \texttt{\{style\}} placeholder is replaced at call time with the dream's pre-assigned style annotation (one of \textit{surreal}, \textit{nightmare}, \textit{peaceful}, \textit{fantastical}), as per appendix A. Each response is validated against a schema requiring exactly four scene objects, each containing a scene number, a short description, an emotion label, a positive SDXL prompt (minimum 10 characters), and a negative SDXL prompt.

\newpage
\hrule
\vskip 0.05in

\begin{quote}\small
You are a dream analyst and visual prompt engineer. The 
user will give you a dream description. Your job:

\begin{enumerate}\itemsep0pt
  \item Split the dream into exactly 4 chronological scenes.
  \item For each scene, write a detailed Stable Diffusion 
  prompt in English.
  \item Apply the style ``\{style\}'' to every prompt.
  \item Identify the dominant emotion of each scene.
\end{enumerate}

Rules for the prompts:
\begin{itemize}\itemsep0pt
  \item Each prompt must include the dreamer as a visible 
  character doing something specific (walking, reaching, 
  swimming, etc.)
  \item Describe body language, pose, and interaction with 
  the environment.
  \item Describe colours, lighting, and atmosphere.
  \item Add ``dreamlike, ethereal, magical'' to every prompt.
  \item Keep prompts between 40 and 70 words.
  \item Put ``unpopulated, empty scenery, no characters'' 
  in all negative prompts.
\end{itemize}

Respond with valid JSON only --- no markdown, no explanation.
\end{quote}
\vskip 0.05in
\hrule

%% file: appendixC.tex
\section*{Appendix C:\\Prompt enrichment details}

SDXL allows both a positive and a negative prompt as inputs. 
Each enriched positive prompt in the DSV pipeline is constructed as:

\vskip 0.1in
\hrule
\vskip 0.05in
\par\noindent
a cohesive cinematic illustration in a consistent 
visual style, [LLM prompt], [style-positive 
tokens], [colour palette], [emotion tokens], masterpiece, best quality, highly detailed, 8k 
resolution, cinematic composition
\vskip 0.05in
\hrule
\vskip 0.1in
\par\noindent
Style-positive tokens and colour palette tokens are taken from this lookup table:
\vskip 0.1in

\begin{center}
\begin{tabular}{p{1.4cm}p{5.8cm}}
\hline
\textbf{Style} & \textbf{Positive tokens; colour palette} \\
\midrule \hline
nightmare & dark horror, ominous shadows, twisted 
forms, unsettling atmosphere, eerie fog, gothic, sinister; 
dark red and black tones, desaturated, high contrast \\
\addlinespace
fantastical & fantasy, magical particles, enchanted, 
mythical creatures, glowing runes, epic scale, vibrant 
colors; rich jewel tones, emerald green, sapphire blue, 
golden highlights \\
\addlinespace
surreal & surreal, dreamlike, floating elements, 
impossible geometry, melting shapes, ethereal lighting, 
fantasy atmosphere; purple and blue tones, iridescent, 
chromatic aberration \\
\addlinespace
peaceful & serene, calm, gentle light, soft focus, 
warm glow, tranquil, harmonious, pastel colors; warm golden 
and soft pink tones, low saturation, gentle gradients \\
\hline
\end{tabular}
\end{center}

\newpage
The emotion tokens are taken from this lookup table:
\vskip 0.1in

\begin{center}
\begin{tabular}{p{1.8cm}p{5.3cm}}
\hline
\textbf{Emotion} & \textbf{Tokens} \\
\hline
mystery & mysterious atmosphere, hidden depths, subtle 
shadows, enigmatic \\
serenity & peaceful, calm water, gentle breeze, soft 
ambient light \\
wonder & awe-inspiring, magical discovery, wide-eyed 
amazement, sparkling \\
amazement & breathtaking vista, overwhelming beauty, 
spectacular, grand reveal \\
fear & creeping dread, looming danger, cold sweat, dark 
corners \\
joy & radiant happiness, warm sunlight, blooming flowers, 
celebration \\
sadness & melancholic, rain, fading colors, lonely 
atmosphere, grey skies \\
anger & intense red, stormy, crashing waves, volcanic, 
fierce energy \\
anxiety & distorted perspective, claustrophobic, uneasy 
lighting, tight spaces \\
awe & vast scale, cosmic, luminous, transcendent beauty \\
nostalgia & golden hour, faded photograph, soft grain, 
warm memories \\
curiosity & inviting path, glowing doorway, beckoning 
light, unknown territory \\
\hline
\end{tabular}
\end{center}
\vskip 0.1in
The negative prompt for SDXL is constructed as:

\vskip 0.1in
\hrule
\vskip 0.05in
\par\noindent
[style-negative tokens],
ugly, blurry, low quality, text, watermark, 
logo, deformed, disfigured, bad anatomy, extra limbs, 
jpeg artifacts, poorly drawn
\vskip 0.05in
\hrule
\vskip 0.1in
\par\noindent
The style-negative tokens for the SDXL negative prompt are taken from this lookup table:
\vskip 0.1in
\begin{center}
\begin{tabular}{p{1.4cm}p{5.8cm}}
\hline
\textbf{Style} & \textbf{Negative tokens} \\
\midrule \hline
nightmare & bright, happy, 
cheerful, cute, cartoon \\
\addlinespace
fantastical & realistic, modern, urban, mundane, dull \\
\addlinespace
surreal & realistic, photograph, mundane, 
ordinary \\
\addlinespace
peaceful & dark, 
scary, violent, chaotic, harsh \\
\hline
\end{tabular}
\end{center}

%% file: appendixD.tex
\section*{Appendix D:\\Vision-language Judge Prompts}

The judge we implemented uses the Qwen2-VL(7B) vision-language model. The first call rates each generated image individually on four per-panel dimensions, using the prompt below, with placeholders \texttt{\{description\}} and \texttt{\{emotion\}} replaced with the panel's description and emotion label.
\vskip 0.1in
\hrule
\vskip 0.05in
\begin{quote}\small
You are evaluating an AI-generated dream visualisation.

Scene description: ``\{description\}''\\
Intended emotion: \{emotion\}

Rate this image on a 1--5 scale for each criterion. 
Respond only with valid JSON.

Guide:
\begin{itemize}\itemsep0pt
  \item \textbf{dreamlike\_quality}: Does it feel 
  surreal and dreamlike? (1 = mundane, 5 = deeply 
  dreamlike)
  \item \textbf{text\_faithfulness}: Does it match the 
  description? (1 = unrelated, 5 = exact match)
  \item \textbf{aesthetic\_quality}: Is it visually 
  compelling? (1 = ugly, 5 = beautiful)
  \item \textbf{character\_presence}: Is a human figure 
  visible doing something? (1 = no person, 5 = clear 
  action)
\end{itemize}
\end{quote}
\vskip 0.05in
\hrule
\vskip 0.1in

The second call rates the full sequence on three per-sequence dimensions, with all four images shown together as input alongside the dream text and individual panel descriptions. The prompt for this is as follows:

\vskip 0.1in
\hrule
\vskip 0.05in
\begin{quote}\small
You are evaluating narrative coherence across 4 dream 
scene images shown above.

The dream: ``\{dream\_text\}''\\
Scene 1: \{s1\} \\
Scene 2: \{s2\} \\
Scene 3: \{s3\} \\
Scene 4: \{s4\}

Rate the 4-image sequence on a 1--5 scale. Respond only 
with valid JSON.

\begin{itemize}\itemsep0pt
  \item \textbf{narrative\_coherence}: Do these tell a 
  visual story? (1 = random, 5 = clear arc)
  \item \textbf{style\_consistency}: Same artistic style 
  throughout? (1 = different, 5 = unified)
  \item \textbf{emotional\_progression}: Does emotion 
  evolve across scenes? (1 = flat, 5 = clear arc)
\end{itemize}
\end{quote}
\vskip 0.05in
\hrule

%% file: appendixE.tex
\section*{Appendix E:\\CLIP Failure Modes}

\vskip 0.1in
\begin{center}
\includegraphics[width=\columnwidth]{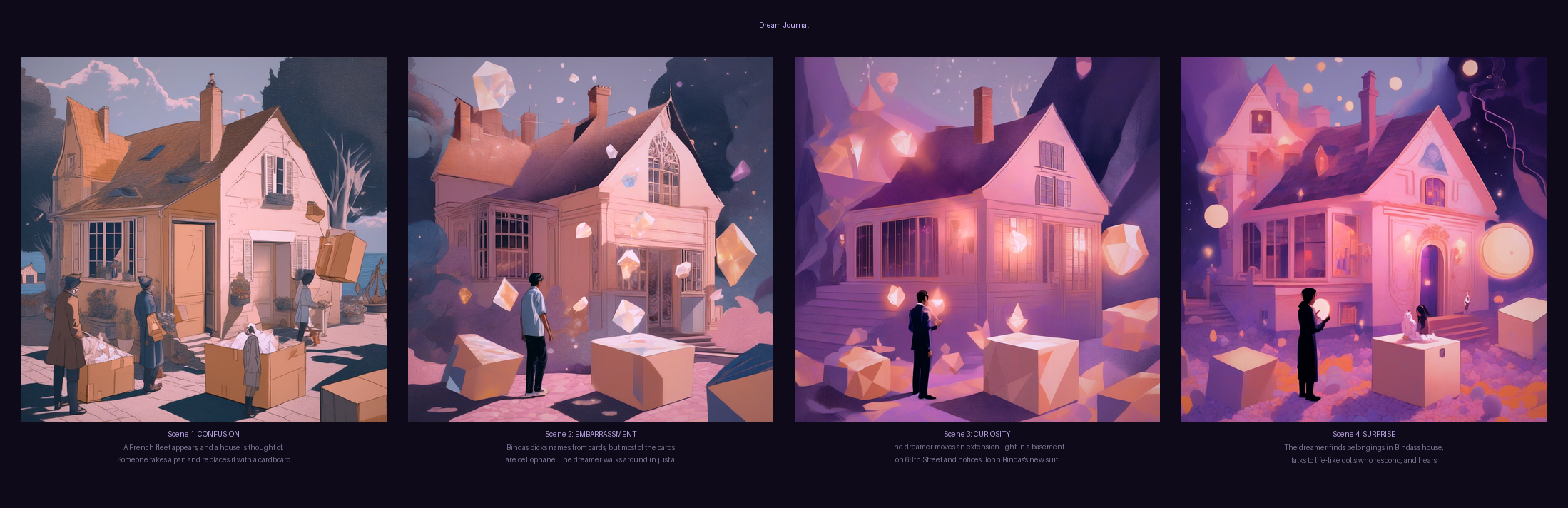}
\end{center}
\textbf{False positive:} (\texttt{DB-3c0442ac}, surreal). 
CLIP~$=0.263$ (above mean) but judge text faithfulness 2.5/5, character presence 3.0/5, emotional progression 1/5. Coherent palette masks character drift across panels.

\vskip 0.1in
\begin{center}
\includegraphics[width=\columnwidth]{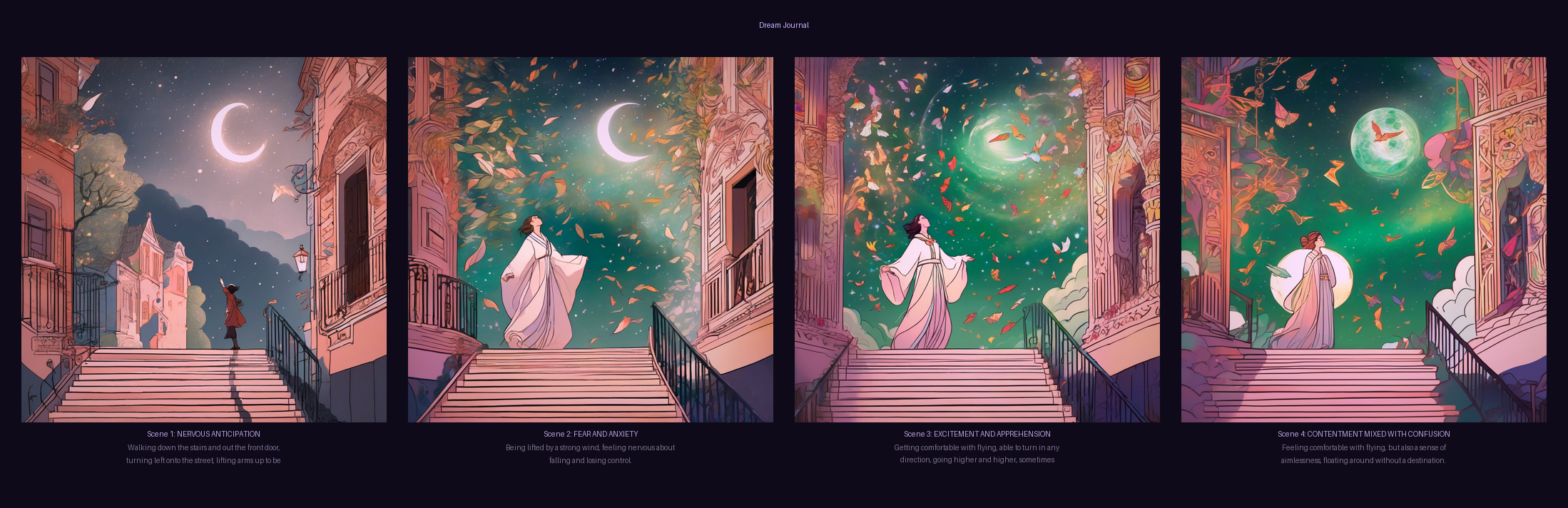}
\end{center}
\textbf{False negative:} (\texttt{DB-dad219b7}, fantastical). CLIP~$=0.231$ (below mean) but judge  faithfulness is 4.25/5, aesthetic quality 5.0/5, style consistency 5/5. Stylisation penalised by CLIP, despite strong semantic alignment.

%% file: appendixF.tex
\section*{Appendix F:\\Additional Dream Outputs}

Below are additional dream sequences spanning three 
style categories, selected to include both high- and low-CLIP score cases. Each entry shows the input dream text, the four generated panels, and the full evaluation signals (CLIP, DINOv2 adjacent coherence, and seven-dimension judge scores reported per panel and per sequence). We provide a subjective commentary for each.

\subsection*{DreamBank Dream DB-3a3aa5d7\\nightmare, 171 words, complexity 4}

\textit{``I was working for a firm. Accidentally I found 
the firm's records that they were carrying on illegal 
activities plus their surface business. Somehow an official 
of the firm learned of my knowledge and held me in 
captivity. After a while in captivity I was able to get 
word to an outside person. However, this didn't have any 
results. But after a while I escaped. As soon as I got 
away, I went to the police and told them my story. Then 
the police and I went to this building and alerted all 
the innocent people who were working for other firms. The 
police then staged a raid, but still the criminals 
wouldn't give up. Then the police used tear gas, and the 
criminals finally gave up. Then there seemed to be some 
time elapsed, and the next thing I can remember is a 
photographer taking my picture in front of the building, 
and the picture was to be printed in the next issue of 
the newspaper. Then I woke up.''}

\vspace{5pt}
\begin{center}
\includegraphics[width=\columnwidth]{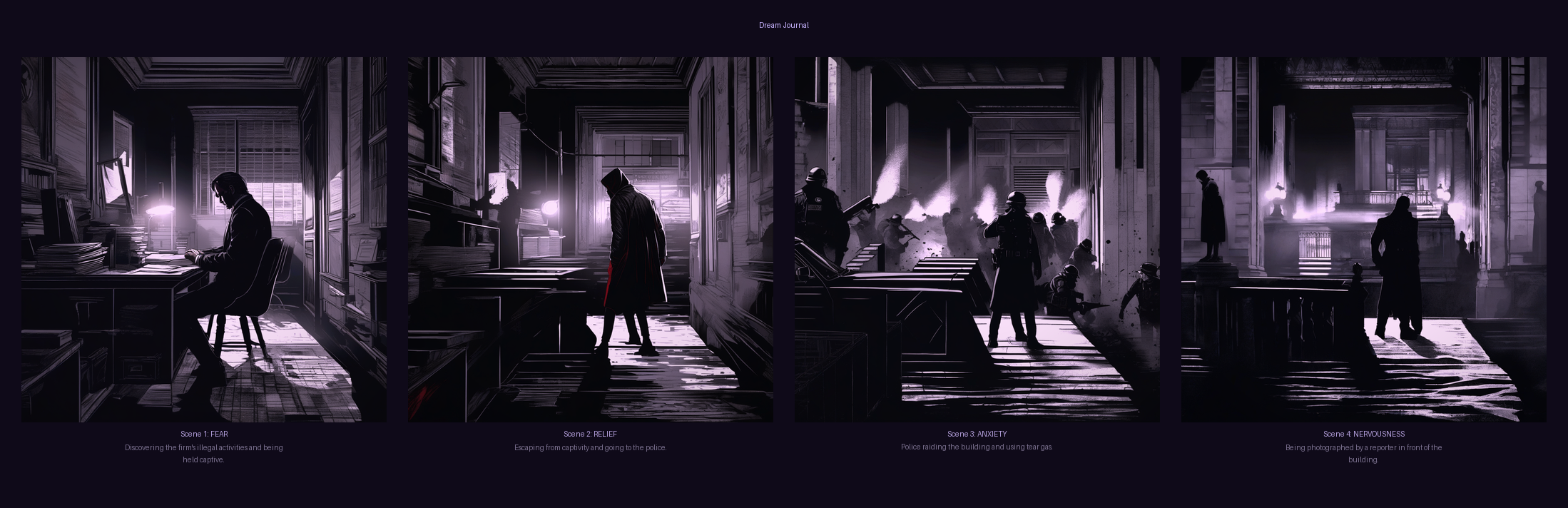}
\end{center}

\vspace{5pt}
\noindent
\textbf{Evaluation.} CLIP: 0.234; DINO: 0.450; Judge per-panel avg. (aesthetic: 4.00, character: 4.00, dreamlike: 3.25, faithfulness: 3.00); Judge per-sequence: (style: 5, narrative: 3, emotional: 3).

\noindent\textbf{Commentary.} The four panels render 
the main sequence: office, escape, police raid, 
exterior, with recognisable elements like helmets 
and tear gas haze in panel 3. The monochromatic palette 
is consistent with the nightmare label, although the 
captivity stage and the final newspaper photograph are 
visualised more loosely.

\vspace{2pt}
\subsection*{DreamBank Dream DB-29e615cd\\fantastical, 194 words, complexity 3}

\textit{``A very small plane came straight down through 
the trees of Lang Lawn. It was spring and a lot of girls 
were watching it come down. Finally it came down with no 
damage resulting. Barbara Ayres, a sophomore here, climbed 
out. Everyone was quite surprised to see that it was she. 
Babs left the plane on the lawn and went into the dorm. 
Joe Warner, a senior, appeared and claimed that the plane 
was his. Nobody seemed to question why Babs was flying 
it. The plane was the smallest one I've ever seen. There 
were two seats, one behind the other, and an average 
individual could just fit into it. Lee Ellsworth, another 
senior, told the onlookers that it took 9.4 seconds to 
get off the ground if Joe did it. Suddenly I was in the 
driver's seat and Joe asked me if I knew where the 
throttle was. It was under a rubber mat on the floor. The 
instrument panel had very little on it. But I got it 
started in 9.0 seconds. I went off the ground and Lee was 
yelling at Joe that I beat his take-off time. Then I 
landed.''}

\vspace{5pt}
\begin{center}
\includegraphics[width=\columnwidth]{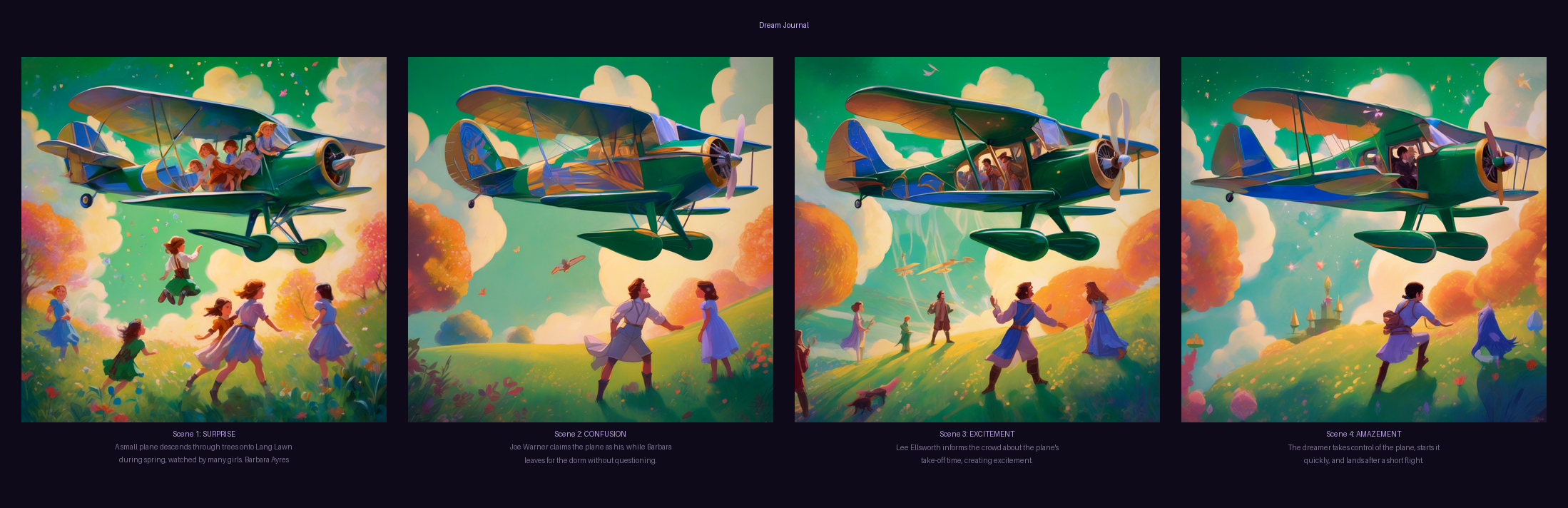}
\end{center}

\vspace{5pt}
\noindent\textbf{Evaluation.} CLIP: 0.257; DINO: 0.932; Judge per-panel avg. (aesthetic: 4.75, character: 4.50, dreamlike: 4.25, faithfulness: 4.00); Judge per-sequence: (style: 5, narrative: 1, emotional: 1).

\noindent\textbf{Commentary.} The fantastical aesthetic is rendered with consistent confidence: pastel skies, autumn foliage, and a recurring biplane. However, the plane remains airborne in all four panels rather than landing after being flown by the dreamer.

\subsection*{DreamBank Dream DB-1c6bab2b\\peaceful, 73 words, complexity 2}

\textit{``I was crossing a valley (a small piece of land 
between two hills), and I heard a friend of mine, a young 
boy in my neighborhood of about the same age as myself, 
shouting to me from the top of one of the hills. I looked 
at him, and he kept on shouting my name and something 
else to me. I started to run very fast until I came home. 
I then awoke.''}

\vspace{5pt}
\begin{center}
\includegraphics[width=\columnwidth]{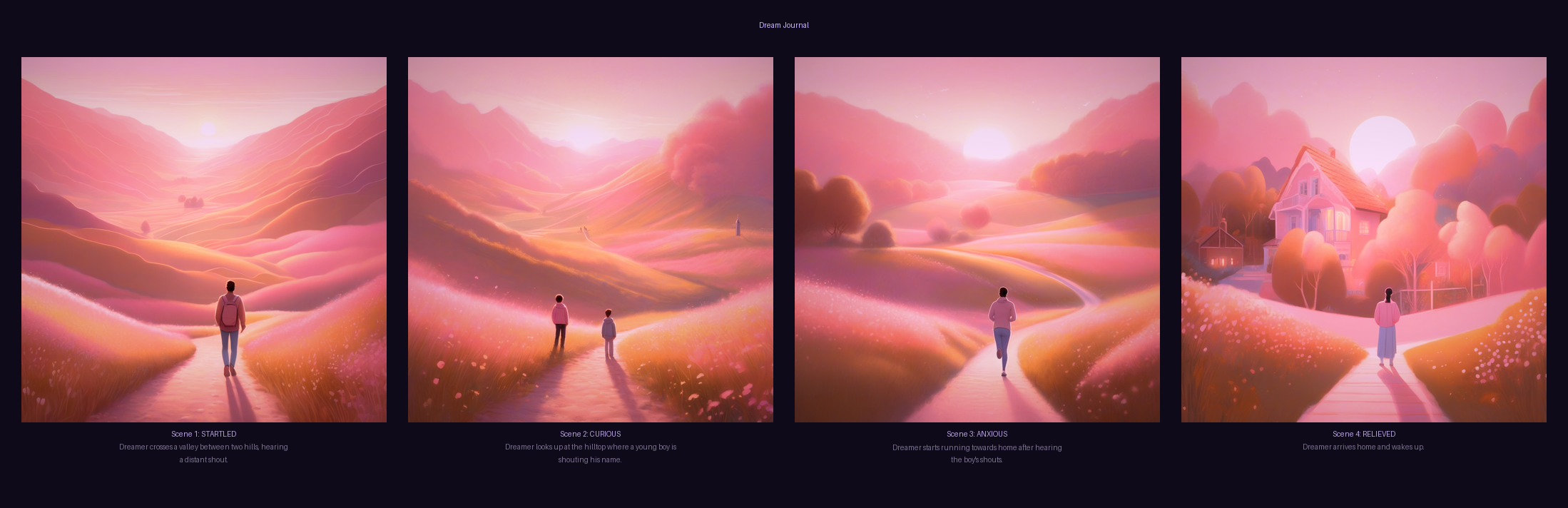}
\end{center}

\vspace{5pt}
\noindent\textbf{Evaluation.} CLIP: 0.267; DINO: 0.709; Judge per-panel avg. (aesthetic: 5.00, character: 4.50, dreamlike: 4.00, faithfulness: 3.75); Judge per-sequence: (style: 5, narrative: 5, 
emotional: 5).

\noindent\textbf{Commentary.} The peaceful aesthetic is 
captured throughout, with a soft pink-and-gold palette 
across valley, path, and home, and the narrative arc 
is largely preserved. The main deviation is in panel 2, 
where the boy stands beside the dreamer on the path 
rather than calling from the top of a hill.

%% file: main.bbl
\begin{thebibliography}{}

\bibitem[\protect\citeauthoryear{Cheatley et~al.}{2020}]{cheatley2020songwriting}
Cheatley, L.; Ackerman, M.; Pease, A.; and Moncur, W. 2020.
\newblock Co-creative songwriting for bereavement support.
\newblock In {\em Proceedings of the 11th International Conference on Computational Creativity (ICCC)}, 33--40.

\bibitem[\protect\citeauthoryear{Domhoff and Schneider}{2008}]{domhoff2008dreambank}
Domhoff, G.~W., and Schneider, A. 2008.
\newblock Studying dream content using the archive and search engine on {DreamBank.net}.
\newblock {\em Consciousness and Cognition} 17(4):1238--1247.

\bibitem[\protect\citeauthoryear{Franceschelli and Musolesi}{2024}]{franceschelli2024creative}
Franceschelli, G., and Musolesi, M. 2024.
\newblock Creative beam search: {LLM}-as-a-judge for improving response generation.
\newblock In {\em Proceedings of the 15th International Conference on Computational Creativity (ICCC)}, 364--368.

\bibitem[\protect\citeauthoryear{Fu et~al.}{2025}]{fMRI}
Fu, Y.; Gao, J.; Yang, B; and Feng, J. 2025.
\newblock Making your dreams a reality: Decoding the dreams into a coherent video story from fMRI signals.
\newblock {\em arXiv preprint arXiv:2501.09350}.

\bibitem[\protect\citeauthoryear{Hall and {Van de Castle}}{1966}]{halldream1966}
Hall, C.~S., and Van~de Castle, R.~L. 1966.
\newblock {\em The Content Analysis of Dreams}.
\newblock New York: Appleton-Century-Crofts.

\bibitem[\protect\citeauthoryear{He et~al.}{2024}]{he2024dreamstory}
He, H.; Yang, H.; Tuo, Z.; Zhou, Y.; Wang, Q.; Zhang, Y.; Liu, Z.; Huang, W.; Chao, H.; and Yin, J. 2024.
\newblock {DreamStory}: Open-domain story visualization by {LLM}-guided multi-subject consistent diffusion.
\newblock {\em IEEE Transactions on Pattern Analysis and Machine Intelligence, 47(12)}.

\bibitem[\protect\citeauthoryear{Hu et~al.}{2021}]{hu2021lora}
Hu, E.~J.; Shen, Y.; Wallis, P.; Allen-Zhu, Z.; Li, Y.; Wang, S.; Wang, L.; and Chen, W. 2021.
\newblock {LoRA}: Low-rank adaptation of large language models.
\newblock {\em arXiv preprint arXiv:2106.09685}.

\bibitem[\protect\citeauthoryear{Krakow et~al.}{2001}]{krakow2001imagery}
Krakow, B.; Hollifield, M.; Johnston, L.; Koss, M.; Schrader, R.; Warner, T.~D.; Tandberg, D.; Lauriello, J.; McBride, L.; Cutchen, L.; Cheng, D.; Emmons, S.; Germain, A.; Melendrez, D.; Sandoval, D.; and Prince, H. 2001.
\newblock Imagery rehearsal therapy for chronic nightmares in sexual assault survivors with post-traumatic stress disorder: A randomized controlled trial.
\newblock {\em JAMA} 286(5):537--545.

\bibitem[\protect\citeauthoryear{Liu et~al.}{2024}]{liu2024dreamllm3d}
Liu, P.; Maverick, K.; Steinmaurer, A.; Picard-Deland, C.; Carr, M.; and Kitson, A. 2024.
\newblock {DreamLLM-3D}: Affective dream reliving using large language model and {3D} generative {AI}.
\newblock In {\em Advances in Neural Information Processing Systems (NeurIPS)}.

\bibitem[\protect\citeauthoryear{Maharana, Hannan, and Bansal}{2022}]{maharana2022storydalle}
Maharana, A.; Hannan, D.; and Bansal, M. 2022.
\newblock {StoryDALL-E}: Adapting pretrained text-to-image transformers for story continuation.
\newblock In {\em European Conference on Computer Vision (ECCV)}.

\bibitem[\protect\citeauthoryear{Oquab et~al.}{2024}]{oquab2024dinov2}
Oquab, M.; Darcet, T.; Moutakanni, T.; Vo, H.; Szafraniec, M.; Khalidov, V.; Fernandez, P.; Haziza, D.; Massa, F.; El-Nouby, A.; Assran, M.; Ballas, N.; Galuba, W.; Howes, R.; Huang, P.-Y.; Li, S.-W.; Misra, I.; Rabbat, M.; Sharma, V.; Synnaeve, G.; Xu, H.; Jegou, H.; Mairal, J.; Labatut, P.; Joulin, A.; and Bojanowski, P. 2024.
\newblock {DINOv2}: Learning robust visual features without supervision.
\newblock {\em Transactions on Machine Learning Research}.

\bibitem[\protect\citeauthoryear{Pan et~al.}{2024}]{pan2023arldm}
Pan, X.; Qin, P.; Li, Y.; Xue, H.; and Chen, W. 2024.
\newblock Synthesizing coherent story with auto-regressive latent diffusion models.
\newblock In {\em IEEE/CVF Winter Conference on Applications of Computer Vision (WACV)}.

\bibitem[\protect\citeauthoryear{Podell et~al.}{2023}]{podell2023sdxl}
Podell, D.; English, Z.; Lacey, K.; Blattmann, A.; Dockhorn, T.; M{\"u}ller, J.; Penna, J.; and Rombach, R. 2023.
\newblock {SDXL}: Improving latent diffusion models for high-resolution image synthesis.
\newblock {\em arXiv preprint arXiv:2307.01952}.

\bibitem[\protect\citeauthoryear{Radford et~al.}{2021}]{radford2021clip}
Radford, A.; Kim, J.~W.; Hallacy, C.; Ramesh, A.; Goh, G.; Agarwal, S.; Sastry, G.; Askell, A.; Mishkin, P.; Clark, J.; Krueger, G.; and Sutskever, I. 2021.
\newblock Learning transferable visual models from natural language supervision.
\newblock In {\em International Conference on Machine Learning (ICML)}, 8748--8763.

\bibitem[\protect\citeauthoryear{Schredl}{2002}]{schredl2002dreamdiary}
Schredl, M. 2002.
\newblock Questionnaires and diaries as research instruments in dream research: Methodological issues.
\newblock {\em Dreaming} 12(1):17--26.

\bibitem[\protect\citeauthoryear{Wang et~al.}{2024}]{wang2024qwen2vl}
Wang, P.; Bai, S.; Tan, S.; Wang, S.; Fan, Z.; Bai, J.; Chen, K.; and others. 2024.
\newblock {Qwen2-VL}: Enhancing vision-language model's perception of the world at any resolution.
\newblock {\em arXiv preprint arXiv:2409.12191}.

\bibitem[\protect\citeauthoryear{Yang et~al.}{2024}]{yang2024qwen25}
Yang, A.; Yang, B.; Zhang, B.; Hui, B.; Zheng, B.; Yu, B.; and others. 2024.
\newblock {Qwen2.5} technical report.
\newblock {\em arXiv preprint arXiv:2412.15115}.

\bibitem[\protect\citeauthoryear{Zheng et~al.}{2024}]{zheng2024mtbench}
Zheng, L.; Chiang, W.-L.; Sheng, Y.; Zhuang, S.; Wu, Z.; Zhuang, Y.; Lin, Z.; Li, Z.; Li, D.; Xing, E.~P.; Zhang, H.; Gonzalez, J.~E.; and Stoica, I. 2024.
\newblock Judging {LLM}-as-a-judge with {MT-Bench} and Chatbot Arena.
\newblock In {\em Advances in Neural Information Processing Systems (NeurIPS)}.

\end{thebibliography}
